\documentclass{article}
\usepackage{iclr2027_conference,times}
\usepackage[T1]{fontenc}
\usepackage[utf8]{inputenc}
\usepackage{amsmath,amssymb,booktabs,array,tabularx}
\usepackage{graphicx}
\usepackage{xcolor,multirow}
\usepackage{subcaption}
\usepackage{float}
\usepackage{tikz}
\usetikzlibrary{arrows.meta,positioning,calc}
\usepackage[section]{placeins}
\usepackage{url}
\usepackage{hyperref}
\hypersetup{hidelinks}
\newcommand{\Q}{\operatorname{Q}}

\newcommand{\KEEP}{\textsc{Keep}}
\newcommand{\SUPPRESS}{\textsc{Suppress}}
\newcommand{\SHARPEN}{\textsc{Sharpen}}

\newcommand{\yes}{\textcolor[rgb]{0,.48,.29}{\ensuremath{\checkmark}}}
\newcommand{\no}{\textcolor[rgb]{.70,.13,.08}{\ensuremath{\times}}}

\newcolumntype{L}[1]{>{\raggedright\arraybackslash}p{#1}}
\title{Anomaly-Guided Observation and Two-Level Semantic Feedback for Training-Free Video Anomaly Analysis}

\author{
  Hangil Park\thanks{Equal contribution.},\enspace 
  Seungheon Kim\footnotemark[1],\enspace
  Tae-Kyun Kim\thanks{Corresponding author.}\\
  Korea Advanced Institute of Science and Technology (KAIST)\\
  \texttt{\{hangil.park, shkseiji5500, kimtaekyun\}@kaist.ac.kr}
 }
 
\iclrfinalcopy 

\begin{document}
\raggedbottom
\maketitle
\lhead{Preprint}
\begin{abstract}
Training-free language-based video anomaly detection (VAD) uses frozen
vision--language models (VLMs) and large language models (LLMs) to localize
abnormal events without task-specific model-weight optimization.
Existing approaches increasingly exploit broader temporal and semantic context,
yet generic context does not necessarily foreground abnormal events or specify
how video-level semantics should modify local anomaly scores.
We introduce a training-free semantic feedback framework that uses
VAD-guided active observation for video anomaly understanding (VAU)
and returns the resulting anomaly-focused description to refine
temporal detection.
Starting from an upstream anomaly-score curve, we adapt reasoning-based
iterative frame selection: high-score regions seed the observations, and
anomaly scores together with VLM relevance assessments guide subsequent
temporal exploration.
A frozen video-language model generates a single open-ended anomaly description
from the selected frames.
The description is returned through two complementary feedback levels:
video-level feedback determines whether the score distribution should be
suppressed, preserved, or sharpened, with selective visual rechecking, while
local feedback retrieves event-related windows and promotes scores only after
target-local visual verification.
All model weights remain frozen, and feedback is applied once
without regenerating the description.
On the HIVAU video-summary evaluations of UCF-Crime and XD-Violence,
VAD-guided observation improves GPT-rated reasonability, detail, and consistency
over the matched CLIP-guided variant.
For VAD, semantic feedback improves the re-evaluated upstream detector
from 84.36\% to 86.25\% AUC on UCF-Crime and from 68.07\% to
72.58\% AP on XD-Violence, where the final AUC is 91.74\%.
\end{abstract}
\section{Introduction}
\label{sec:intro}

Video anomaly detection (VAD) aims to identify and temporally localize
abnormal events in long, untrimmed videos
\citep{sultani2018ucf,wu2020xd}.
Recent language-based approaches use pretrained vision--language models
(VLMs) and large language models (LLMs) to interpret visual observations and
estimate anomaly scores without task-specific model-weight updates
\citep{zanella2024lavad,lin2025urf}.
In this setting, detection depends on the evidence and context supplied
to frozen models at inference time.

Local observations alone can be insufficient for temporally extended or
ambiguous events.
Training-free VAD methods therefore increasingly incorporate broader temporal
and semantic context through caption aggregation, contextual reasoning, and
score refinement
\citep{zanella2024lavad,lim2026corevad,mo2026reci}.
However, broader context is not necessarily \emph{anomaly-focused}.
A generic video summary may devote substantial capacity to common scene content
while representing the abnormal event only indirectly.
Video anomaly understanding (VAU), in contrast, explicitly characterizes
what happened and why it is anomalous
\citep{zhang2025holmes,lin2025urf}.
This suggests a complementary role for VAU: beyond serving as an explanation
output, its anomaly-focused semantics can provide hypotheses for reconsidering
temporal VAD predictions.

Obtaining such semantics from an imperfect detector is itself challenging.
Initial anomaly scores help locate suspicious evidence~\citep{lin2025urf},
but one-shot sampling cannot adapt to evidence discovered during observation.
A.I.R.~\citep{zou2026air} iteratively evaluates frame relevance and reallocates
subsequent observations for video question answering, while Learning to
Watch~\citep{mo2026watch} learns interleaved observation--reasoning policies
for VAU.
We adapt active observation to training-free VAU: the detector
score curve initializes and guides observation, while VLM
relevance assessments obtained during acquisition influence
subsequent temporal exploration.

\begin{figure}[t]
\centering
\begin{subfigure}[t]{0.48\linewidth}
\centering
\includegraphics[width=\linewidth]{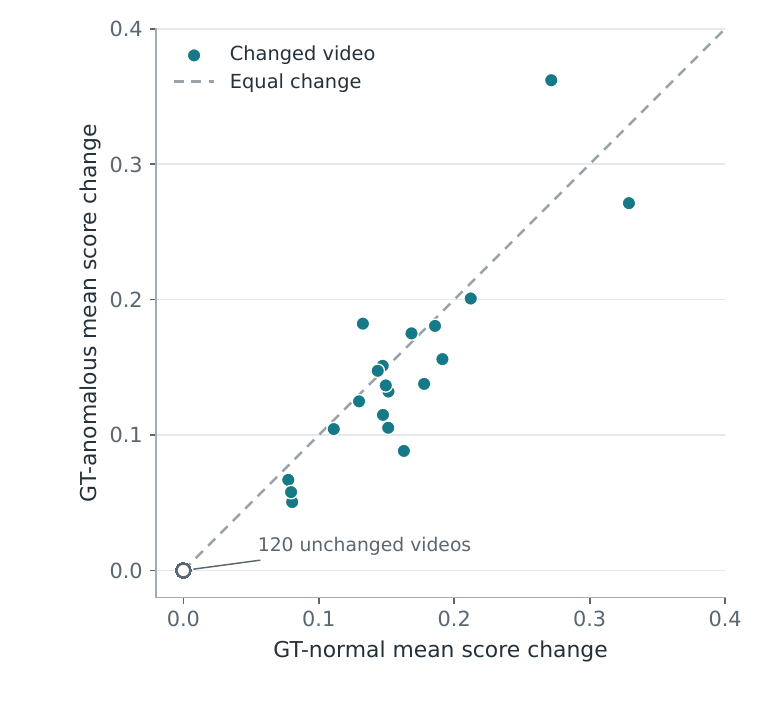}
\vspace{-10mm}
\caption{}
\label{fig:refinement-scatter}
\end{subfigure}\hfill
\begin{subfigure}[t]{0.48\linewidth}
\centering
\includegraphics[width=\linewidth]{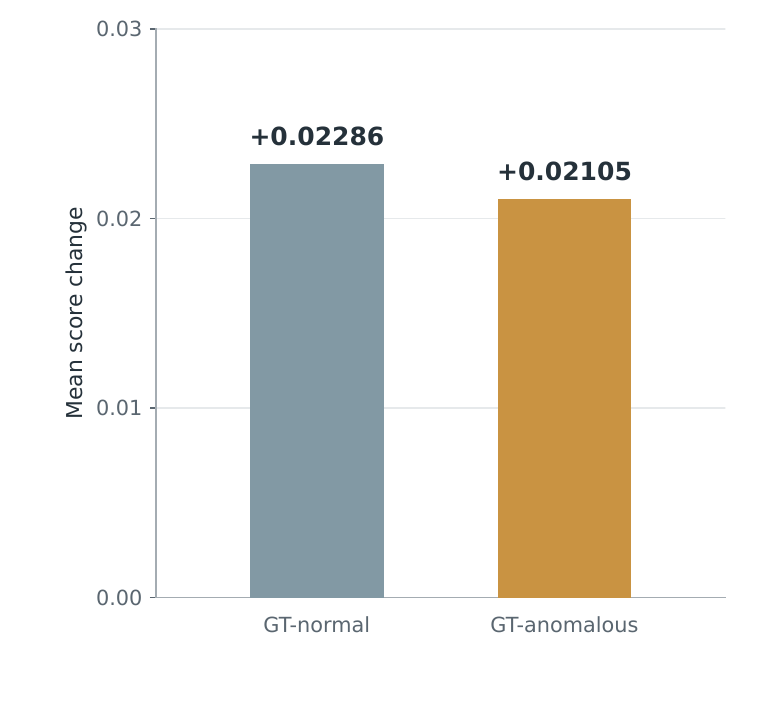}
\vspace{-10mm}
\caption{}
\label{fig:refinement-means}
\end{subfigure}

\caption{Changes from initial (RAW) to URF-HVAA-refined scores on UCF-Crime. (a) Per-video changes in GT-normal and GT-anomalous mean scores; the dashed line denotes equal changes, and 120 unchanged videos overlap at the origin. (b) Mean changes averaged equally across all 140 test videos containing both GT region types.}
\vspace{-5mm}
\label{fig:refinement-motivation}
\end{figure}
Returning the resulting description to VAD raises a second problem.
A video-level judgment that an anomaly occurred does not identify which
temporal regions warrant score promotion, and textual agreement with a local
caption does not establish visible abnormal evidence in that interval.
In our UCF-Crime diagnostic, URF-HVAA refinement produces
similar video-averaged score increases in ground-truth normal
and anomalous regions (Fig.~\ref{fig:refinement-motivation}).
This observation motivates distinguishing \emph{how the video-level
score distribution should be treated} from \emph{where local promotion
is visually warranted}.
Prior work connects detection and understanding through chained anomaly
analysis, training-free semantic grounding, or detector--LVLM learning
\citep{lin2025urf,gao2025vagu,hashimoto2026must}.
We study a specific return mechanism: a completed VAU description guides
a single refinement pass over an existing temporal anomaly-score sequence,
separating video-level modulation from local promotion gated by visual
evidence.

Our framework couples VAD-guided active observation with this two-level
semantic feedback. The description is generated once, evaluated as a VAU
output, and returned to refine detection without regeneration.
Experiments show improved GPT-rated description quality and complementary
detection gains from the two feedback levels, with the local contribution
more apparent in AP than in AUC.

Our main contributions are:
\begin{itemize}
    \item We introduce a training-free semantic feedback framework
    that separates video-level score modulation from local
    refinement supported by visual evidence.

    \item We develop VAD-guided active observation for anomaly-focused
    video understanding, allowing newly acquired visual evidence
    to guide subsequent temporal exploration.

    \item We show that the generated descriptions are useful both
    as anomaly explanations and as feedback for temporal detection,
    evaluating description quality and detection refinement separately.
\end{itemize}
\section{Related Work}
\label{sec:related}
\paragraph{Context-aware training-free VAD.}
LAVAD combines frame captioning, temporal summarization, and score refinement
using video--text similarity~\citep{zanella2024lavad}, while EventVAD performs event-aware
temporal decomposition before MLLM-based anomaly reasoning
~\citep{shao2025eventvad}.
CoReVAD uses a frozen VLM to generate local decisions and descriptions and
then incorporates broader visual--semantic context
~\citep{lim2026corevad}.
ReCI combines video-level semantic context, retrieved reference context, and
confidence-guided temporal refinement~\citep{mo2026reci}.
PANDA introduces agentic planning and knowledge retrieval
~\citep{yang2025panda}, whereas URF-HVAA chains multiple anomaly-oriented
priors through a unified zero-shot analysis pipeline~\citep{lin2025urf}.
We use an anomaly-focused interpretation to guide temporal score refinement.

\paragraph{Anomaly-focused video understanding.}
VAU extends anomaly analysis from temporal scores to semantic descriptions of
abnormal events and their context.
Holmes-VAU combines hierarchical instruction data with anomaly-focused
temporal sampling~\citep{zhang2025holmes}, while VADER incorporates
context-aware sampling and relation-aware representations
~\citep{cheng2026vader}.
URF-HVAA passes temporal, textual, and spatial anomaly priors into final
understanding~\citep{lin2025urf}, and the VAGU benchmark and GtS framework address joint anomaly
grounding and understanding~\citep{gao2025vagu}.
Recent methods additionally make visual observation adaptive.
A.I.R. combines query-conditioned initialization, VLM relevance assessment,
and localized resampling~\citep{zou2026air};
Learning to Watch learns interleaved observation--reasoning trajectories
~\citep{mo2026watch}; and AgenticVAU coordinates exploration and verification
across multiple agents~\citep{duan2026agenticvau}.
We use adaptive observation to construct the description returned to the detector.

\paragraph{VAD--VAU interaction and semantic feedback.}
Detection and understanding can exchange information at different stages of
anomaly analysis.
URF-HVAA propagates intermediate anomaly priors across detection and
understanding stages~\citep{lin2025urf}.
Glance then Scrutinize is a training-free framework that uses semantic
interpretations to guide subsequent temporal grounding
\citep{gao2025vagu}.
MuST-VAD instead forms an alternating learning loop in which detector
predictions guide LVLM adaptation and updated semantic representations are
returned to retrain the detector~\citep{hashimoto2026must}.
Our method targets a different interaction regime: a completed VAU description
is returned to an existing anomaly-score sequence in a single inference-time
pass while all model weights remain frozen.
The return path separates video-level score modulation from
event-conditioned, visually authorized local refinement without treating the
generated description as dense temporal supervision.
\section{Method}
\label{sec:method}
\subsection{Problem setup and overview}
\label{sec:setup}
Given an untrimmed video $V=(v_0,\ldots,v_{T-1})$ of $T$ frames,
where $v_t$ denotes frame $t$, our framework outputs an open-ended
anomaly description $Y$ and a refined temporal anomaly-score sequence.

A frozen upstream analyzer provides local captions $C$, an initial
anomaly-score sequence $b^{(0)}$, a suspicious-activity phrase $h$,
and optional spatial overlays.
In the main configuration, $b^{(0)}$ is the URF-HVAA refined score sequence.
The same scores guide active observation and provide the starting point for
semantic refinement.
The index $i$ ranges over the available score anchors, with $b_i^{(0)}$ anchored at original frame $16i$.
For observation, the scores are smoothed and expanded to frame-level values
as described in Appendix~\ref{app:observation}.

Figure~\ref{fig:overview} summarizes the three components. Score-guided observation constructs visual evidence and a VAU response. The response then supplies a video-level interpretation and local event queries:
\begin{align}
F &= \Pi_{16}\!\left(\mathcal A(V,b^{(0)},q)\right), &
Y &= \mathcal G(\widetilde V_F,q,h), \label{eq:forward}\\
b^{(v)} &= \mathcal R_v(b^{(0)},Y,C,V), &
b^{(1)} &= \mathcal R_\ell(b^{(v)};b^{(0)},Y,C,V). \label{eq:return}
\end{align}
Here $q$ denotes the common VAU instruction, $\mathcal A$ the iterative
observation procedure, and $\mathcal G$ the frozen VAU answerer.
The projection $\Pi_{16}$ produces a chronologically ordered sequence
$F$ of 16 original-frame indices.
The visual input $\widetilde V_F$ contains the corresponding frames
after optional upstream spatial rendering.
The two return operators act at different temporal scales:
$\mathcal R_v$ modifies the video-wide score distribution, whereas
$\mathcal R_\ell$ retrieves candidate windows using $Y$, $C$, and
$b^{(0)}$ and promotes selected anchors in the current scores
after target-local visual verification.
All model weights remain frozen.
The description $Y$ is generated once and is not rewritten after feedback.
\begin{figure}[t]
\centering
\includegraphics[width=\linewidth]{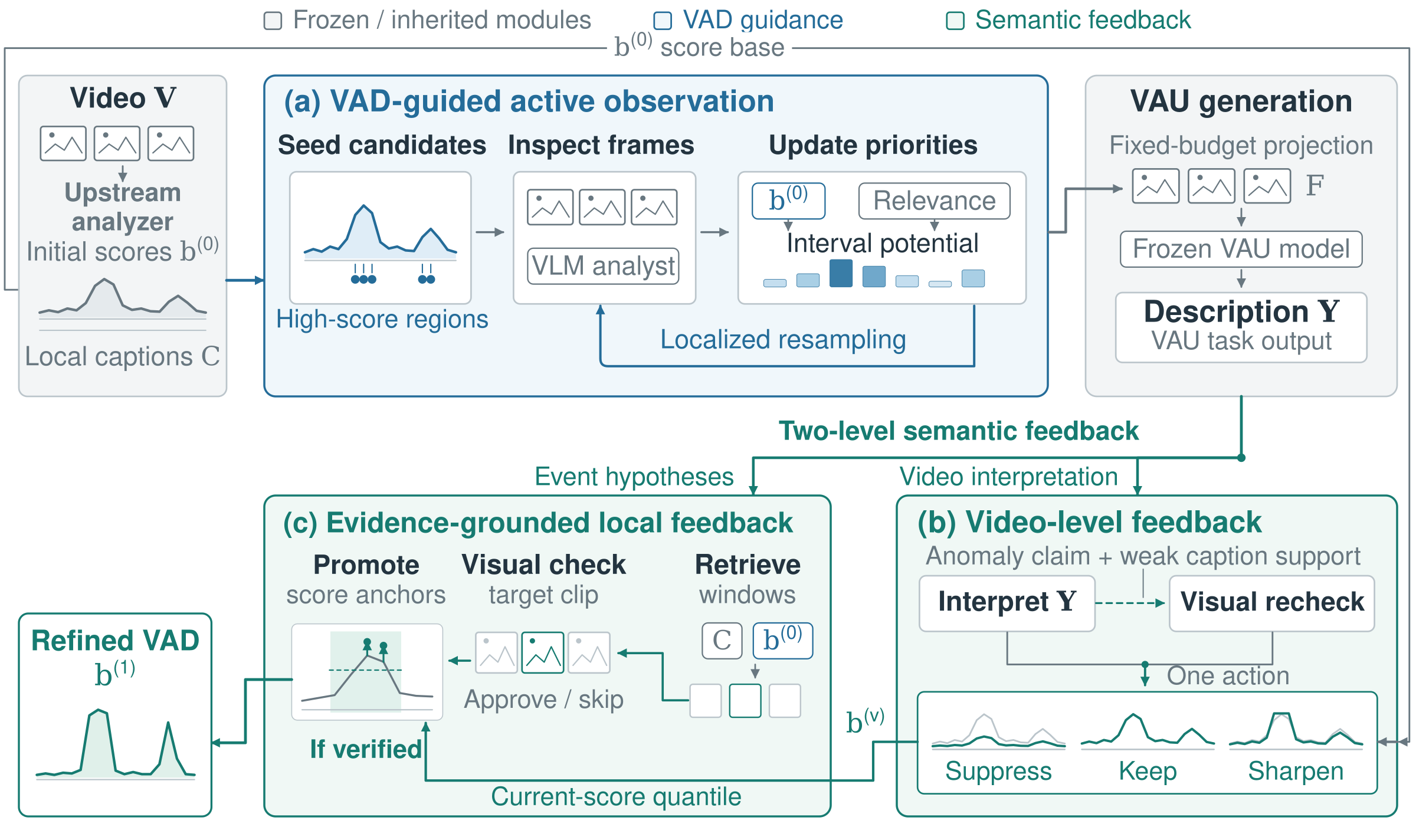}
\caption{Anomaly-guided observation and two-level semantic feedback.
(a) Initial scores $b^{(0)}$ guide seeding and interval priorities within
the inherited A.I.R. observation loop; the frames indexed by $F$ yield one description $Y$.
(b) The video interpretation selects one score modulation, with selective
visual rechecking of anomaly hypotheses.
(c) For anomaly judgments, event hypotheses from $Y$, captions $C$,
and $b^{(0)}$ retrieve windows;
target-clip evidence authorizes promotion, and current-score quantiles
select the anchors.
The score path is $b^{(0)}\!\rightarrow b^{(v)}\!\rightarrow b^{(1)}$.
All visual models use frames derived from $V$ and remain frozen.
Curves and tokens are schematic; the single return pass precedes final
smoothing and expansion.}
\label{fig:overview}
\end{figure}

\subsection{Score-guided active evidence acquisition for VAU}
\label{sec:observation}

The common VAU instruction asks for anomalous activities without
naming a video-specific event. We therefore use $b^{(0)}$ as a
video-specific prior for locating evidence, replacing query--frame
CLIP similarity~\citep{radford2021clip} in the A.I.R. observation
procedure~\citep{zou2026air}.

High-score regions seed the candidate frames. During subsequent
observation, the inherited interval-potential mechanism combines
anomaly scores with VLM relevance assessments of inspected
candidates. Localized density sampling then proposes further
frames around retained locations. Newly inspected evidence
therefore affects where the procedure looks next.
The resulting selection is projected to the answerer's 16-frame
input. The interval potential and observation settings are given
in Appendix~\ref{app:observation}.

The answerer $\mathcal G$ generates $Y$ from these frames,
retaining URF-HVAA's suspicious-phrase prompting and gated spatial
conditioning~\citep{lin2025urf}.
The same description is evaluated as the VAU output and supplied
to both feedback levels.
Generation settings are provided in
Appendix~\ref{app:generation}.

\subsection{Video-level semantic feedback}
\label{sec:video}
The first feedback level determines how the video's overall score distribution should be treated. It combines a tentative interpretation from $Y$ with selective visual rechecking when that interpretation is weakly reflected in the local captions.

\paragraph{Tentative interpretation and visual rechecking.}
A lexical parser assigns $Y$ to one of four coarse states:
\texttt{normalish}, \texttt{unclear}, \texttt{mixed}, or
\texttt{anomalyish}.
These states induce tentative
$\SUPPRESS$, $\KEEP$, $\SHARPEN$, and $\SHARPEN$ actions, respectively.
The state \texttt{mixed} only denotes simultaneous matches to normality and
anomaly expressions and does not resolve their semantic relationship.

We selectively invoke visual rechecking when the VAU description indicates a
potential anomaly that is weakly supported by the existing local captions.
The frozen verifier observes frames sampled from the full video and from a
high-score interval together with the VAU hypothesis.
The hypothesis is treated as untrusted text: the verifier is instructed to rely
on visible abnormal evidence rather than inferred intent or criminality.
Ambiguous or internally inconsistent verification conservatively preserves the
input scores, whereas clearly unsupported anomaly hypotheses may trigger
suppression; otherwise the tentative action is retained.
Exactly one video-level operation is applied.
The eligibility threshold and visual sampling settings are
provided in Appendix~\ref{app:feedback}.

\paragraph{Score modulation.}
Let $a$ be the final action, $\Q_\tau$ the $\tau$-quantile operator
computed with linear interpolation, and $m\in[0,1]$ the minimum
peak threshold. The peak set is
\begin{equation}
\theta_v=\max\{m,\Q_{0.70}(b^{(0)})\},\qquad
H=\{i:b_i^{(0)}\geq\theta_v\}.\label{eq:peak}
\end{equation}
Let $\alpha,\beta\in[0,1]$ denote the suppression and non-peak
attenuation factors, respectively, and let $f\in[0,1]$ denote the
peak-score floor. Video-level modulation is
\begin{equation}
b_i^{(v)}=\begin{cases}
\alpha b_i^{(0)}, & a=\SUPPRESS,\\
b_i^{(0)}, & a=\KEEP,\\
\max\{b_i^{(0)},f\}, & a=\SHARPEN,\ i\in H,\\
\beta b_i^{(0)}, & a=\SHARPEN,\ i\notin H,
\end{cases}\label{eq:video-update}
\end{equation}
followed by clipping to $[0,1]$.
Suppression reduces the overall score scale. Sharpening attenuates non-peak
scores and imposes a floor on scores in $H$, when this set is nonempty.
These operations modulate the existing score sequence without independently
localizing an event interval.
A single shared parameter configuration is used across all evaluated
datasets and compared variants; its values are reported in Appendix~\ref{app:feedback}.
The transformed scores are not interpreted as calibrated anomaly probabilities.

\subsection{Evidence-grounded local feedback}
\label{sec:local}

Local feedback separates three decisions: event semantics proposes
candidate windows, visual verification authorizes promotion,
and current-score ranks select the anchors to update.

\paragraph{Event-conditioned retrieval.}
A separate parser extracts visible event hypotheses from $Y$.
Local promotion is enabled only when this parser judges the video
anomalous, independently of the video-level action.
TF--IDF matching against $C$ retrieves non-overlapping caption
windows for each hypothesis, with $b^{(0)}$ providing additional
ranking and filtering cues.

\paragraph{Target-local visual verification.}
A verifier inspects each candidate window and surrounding frames,
using the event hypothesis and matched captions as noisy retrieval
cues. Promotion requires the verifier to identify direct abnormal
evidence inside the target clip; surrounding context alone cannot
authorize it. The verifier authorizes a window, while anchor
selection remains score-based.

\paragraph{Local score promotion.}
Initialize the current score sequence $x=b^{(v)}$.
For an approved window $w$ in original-frame coordinates containing
at least one score anchor, let $x_w=(x_i:16i\in w)$.
Let $\tau_\ell\in[0,1]$ be the local quantile level and
$\lambda\in[0,1]$ the promotion strength:
\begin{equation}
M_w=\{i:16i\in w,\ x_i\geq\Q_{\tau_\ell}(x_w)\},\qquad
x_i\leftarrow x_i+\lambda(1-x_i),\quad i\in M_w.
\label{eq:local-update}
\end{equation}
Other anchors remain unchanged.
Approved windows are processed sequentially, recomputing the
quantile from the current scores after preceding updates.
After all updates, set $b^{(1)}=x$.
Feedback settings are given in Appendix~\ref{app:feedback};
final smoothing, frame-level expansion, and evaluation are
described in Appendix~\ref{app:metrics}.
\section{Experiments}
\label{sec:experiments}

\subsection{Experimental Setup}

\paragraph{Datasets and metrics.}
We evaluate video anomaly detection on UCF-Crime, XD-Violence, UBNormal,
and MSAD, following the evaluation protocols used by prior VAD methods
~\citep{sultani2018ucf,wu2020xd,acsintoae2022ubnormal,zhu2024msad}.
For UCF-Crime and UBNormal, the main benchmark metric is frame-level AUC; we additionally report UCF-Crime AP in the feedback analyses.
For XD-Violence and MSAD, we report frame-level AUC and average precision
(AP); MSAD follows its official micro-AUC/AP evaluation protocol.
For VAU, we follow the HIVAU-70K video-summary setting used by URF-HVAA,
with 251 UCF-Crime and 800 XD-Violence videos
~\citep{zhang2025holmes,lin2025urf}.
We report BLEU~\citep{papineni2002bleu},
CIDEr~\citep{vedantam2015cider},
METEOR~\citep{banerjee2005meteor}, and
ROUGE-L~\citep{lin2004rouge}.
We also report GPT-based scores for reasonability, detail,
and consistency following Hawk~\citep{tang2024hawk}.
The GPT evaluator compares generated and reference descriptions without access to video pixels.

\paragraph{Models and implementation.}
All model weights remain frozen throughout inference.
The upstream URF-HVAA system provides local captions and the refined
temporal anomaly-score curve.
Qwen2.5-VL-7B-Instruct~\citep{bai2025qwen25vl} serves as the candidate analyst during active
observation, and VideoLLaMA3-7B~\citep{zhang2025videollama3} generates the final open-ended VAU description.
Qwen2-VL-7B-Instruct~\citep{wang2024qwen2vl} is used independently for video-level rechecking and
target-local visual verification.
The final VAU answerer receives 16 selected frames; this is the answer-generation
budget rather than the total number of visual observations made by the complete
pipeline.
Implementation details are given in Appendices~\ref{app:observation}--\ref{app:feedback}; metric definitions and inference costs are reported in Appendix~\ref{app:metrics}.

\paragraph{Main configuration and comparisons.}
The main configuration uses URF-HVAA refined scores for both observation
and feedback initialization, and applies video-level followed by local feedback.
Tables~\ref{tab:vad-main} and~\ref{tab:vau-main} compare this configuration
with existing methods; ablations examine observation strategies and feedback stages.
A single feedback parameter configuration is shared across all evaluated
datasets and held fixed across the compared variants.

\subsection{Video Anomaly Detection}
\label{sec:vad-results}

\begin{table}[!t]
\centering
\caption{
Comparison with representative video anomaly detection methods.
The upper block includes methods using additional training or reference data.
The lower block reports selected zero-shot, training-free methods.
Legacy comparison values follow the published VAD comparison in
URF-HVAA~\citep{lin2025urf}; MSAD results for RTFM and MGFN follow the
official MSAD benchmark~\citep{zhu2024msad}.
Bold denotes the highest reported value within the lower block.
All values are percentages; ``--'' denotes an unreported metric.
}
\label{tab:vad-main}

\scriptsize
\setlength{\tabcolsep}{2.8pt}
\renewcommand{\arraystretch}{1.07}

\resizebox{\linewidth}{!}{%
\begin{tabular}{@{}lcc|c|cc|c|cc@{}}
\toprule
\multirow{2}{*}{Method}
& \multirow{2}{*}{Zero-shot}
& \multirow{2}{*}{Training-free}
& UCF-Crime
& \multicolumn{2}{c|}{XD-Violence}
& UBNormal
& \multicolumn{2}{c}{MSAD} \\
\cmidrule(lr){4-4}
\cmidrule(lr){5-6}
\cmidrule(lr){7-7}
\cmidrule(l){8-9}
& & &
AUC(\%)
& AUC(\%) & AP(\%)
& AUC(\%)
& AUC(\%) & AP(\%) \\
\midrule

\multicolumn{9}{@{}l}{\textit{Additional training or reference data}} \\

RTFM~\citep{tian2021rtfm}
& $\times$ & $\times$
& 83.31
& -- & 77.81
& 60.94
& 86.7 & 66.3 \\

MGFN~\citep{chen2023mgfn}
& $\times$ & $\times$
& 86.98
& -- & 80.11
& --
& 85.0 & 63.5 \\

MULDE~\citep{micorek2024mulde}
& $\times$ & $\times$
& 78.50
& -- & --
& 72.80
& -- & -- \\

EGO~\citep{ding2025ego}
& $\times$ & $\times$
& 81.71
& -- & 65.77
& --
& 87.3 & 64.4 \\

AnomalyRuler~\citep{yang2024anomalyruler}
& $\times$ & $\checkmark$
& --
& -- & --
& 71.90
& -- & -- \\

\midrule

\multicolumn{9}{@{}l}{\textit{Zero-shot, training-free}} \\

LAVAD~\citep{zanella2024lavad}
& $\checkmark$ & $\checkmark$
& 80.28
& 85.36 & 62.01
& --
& -- & -- \\

EventVAD~\citep{shao2025eventvad}
& $\checkmark$ & $\checkmark$
& 82.03
& 87.51 & 64.04
& --
& -- & -- \\

URF-HVAA~\citep{lin2025urf}
& $\checkmark$ & $\checkmark$
& 84.28
& 91.34 & 68.07
& \textbf{68.98}
& 85.9 & 76.4 \\

CoReVAD~\citep{lim2026corevad}
& $\checkmark$ & $\checkmark$
& 82.51
& 91.44 & 70.94
& --
& -- & -- \\

ReCI~\citep{mo2026reci}
& $\checkmark$ & $\checkmark$
& 81.51
& 91.60 & 72.23
& --
& -- & -- \\

\midrule

\textbf{Ours}
& $\checkmark$ & $\checkmark$
& \textbf{86.25}
& \textbf{91.74} & \textbf{72.58}
& 68.67
& \textbf{86.0} & \textbf{76.9} \\

\bottomrule
\end{tabular}%
}
\end{table}

Table~\ref{tab:vad-main} compares the complete framework with zero-shot,
training-free methods and includes trained or reference-based methods for
broader coverage of UBNormal and MSAD.

Our framework obtains 86.25\% AUC on UCF-Crime,
91.74\% AUC / 72.58\% AP on XD-Violence, and
86.0\% AUC / 76.9\% AP on MSAD.
Relative to the published URF-HVAA results, these correspond to gains of
1.97 AUC points on UCF-Crime, 0.40/4.51 AUC/AP points on
XD-Violence, and 0.1/0.5 AUC/AP points on MSAD.
On UBNormal, our method obtains 68.67\% AUC, compared with
68.98\% for URF-HVAA.
These comparisons favor our method on the three real-world benchmarks,
with a small decrease on synthetic UBNormal. Table~\ref{tab:vad-ablation} separately compares the feedback
stages with the upstream scores under the same evaluation setting.

\subsection{Video Anomaly Understanding}
\label{sec:vau-results}

\begin{table}[!htbp]
\centering
\caption{
Video anomaly understanding on the HIVAU video-summary setting.
GPT-R, GPT-D, and GPT-C denote reasonability, detail, and consistency,
respectively.
Published baseline values follow URF-HVAA~\citep{lin2025urf};
the final row reports our VAD-guided active observation pipeline. URF-HVAA (re-eval.) reports the released URF-HVAA descriptions under the
same evaluator used for our generated descriptions.
}
\label{tab:vau-main}
\setlength{\tabcolsep}{1.6pt}
\renewcommand{\arraystretch}{1.16}
\scriptsize

\resizebox{\linewidth}{!}{%
\begin{tabular}{@{}l|cccc|ccc|cccc|ccc@{}}
\toprule
\multirow{2}{*}{Method}
& \multicolumn{7}{c|}{UCF-Crime}
& \multicolumn{7}{c}{XD-Violence}\\
\cmidrule(lr){2-8}\cmidrule(l){9-15}
& BLEU & CIDEr & METEOR & ROUGE-L & GPT-R & GPT-D & GPT-C
& BLEU & CIDEr & METEOR & ROUGE-L & GPT-R & GPT-D & GPT-C\\
\midrule

\multicolumn{15}{@{}l@{}}{%
    \textit{Supervised Fine-tuning}}\\
\addlinespace[2pt]

Hawk~\citep{tang2024hawk}
& 0.379 & 0.008 & 0.217 & 0.187 & 0.255 & 0.580 & 0.214
& 0.375 & 0.016 & 0.176 & 0.188 & 0.408 & 0.586 & 0.365\\

HolmesVAU~\citep{zhang2025holmes}
& 0.435 & 0.021 & 0.194 & 0.257 & 0.448 & 0.356 & 0.391
& 0.376 & 0.011 & 0.182 & 0.253 & 0.715 & 0.581 & 0.673\\

\midrule

\multicolumn{15}{@{}l@{}}{%
    \textit{Zero-shot, training-free}}\\
\addlinespace[2pt]

InternVideo2.5-8B~\citep{wang2025internvideo25}
& 0.159 & 0.011 & 0.088 & 0.103 & 0.240 & 0.266 & 0.205
& 0.209 & 0.013 & 0.119 & 0.130 & 0.456 & 0.447 & 0.433\\

VideoChat-Flash-2B~\citep{li2026videochatflash}
& 0.165 & 0.008 & 0.108 & 0.168 & 0.488 & 0.283 & 0.404
& 0.277 & 0.026 & 0.144 & 0.186 & 0.690 & 0.576 & 0.627\\

VideoLLaMA3-7B~\citep{zhang2025videollama3}
& 0.215 & 0.014 & 0.117 & 0.156 & 0.463 & 0.289 & 0.384
& 0.290 & 0.022 & 0.141 & 0.169 & 0.568 & 0.487 & 0.499\\

URF-HVAA~\citep{lin2025urf}
& 0.345 & \textbf{0.023} & 0.175 & 0.188 & 0.512 & 0.428 & 0.444
& 0.399 & 0.029 & 0.198 & 0.200 & 0.721 & 0.707 & 0.668\\

\midrule
URF-HVAA (re-eval.)
 & 0.345 & \textbf{0.023} & 0.175 & 0.188 & 0.506 & 0.443 & 0.451
 & 0.404 & 0.028 & 0.202 & 0.200 & 0.721 & 0.708 & 0.670\\

\textbf{Ours}
& \textbf{0.386} & \textbf{0.023} & \textbf{0.183} & \textbf{0.196} & \textbf{0.524} & \textbf{0.475} & \textbf{0.466}
& \textbf{0.465} & \textbf{0.047} & \textbf{0.211} & \textbf{0.212} & \textbf{0.731} & \textbf{0.730} & \textbf{0.682}\\

\bottomrule
\end{tabular}%
}
\end{table}

Table~\ref{tab:vau-main} evaluates the generated descriptions before
they are used for detector refinement.
Compared with the re-evaluated URF-HVAA descriptions, our method
improves all three GPT-rated dimensions on both datasets.
Lexical metrics also improve, except for UCF-Crime CIDEr, which
is unchanged at the reported precision.
Table~\ref{tab:vau-ablation} further examines the effects of
anomaly guidance and active observation.

\paragraph{Qualitative analysis.}
\begin{figure}[t]
    \centering
    \includegraphics[width=.99\linewidth]{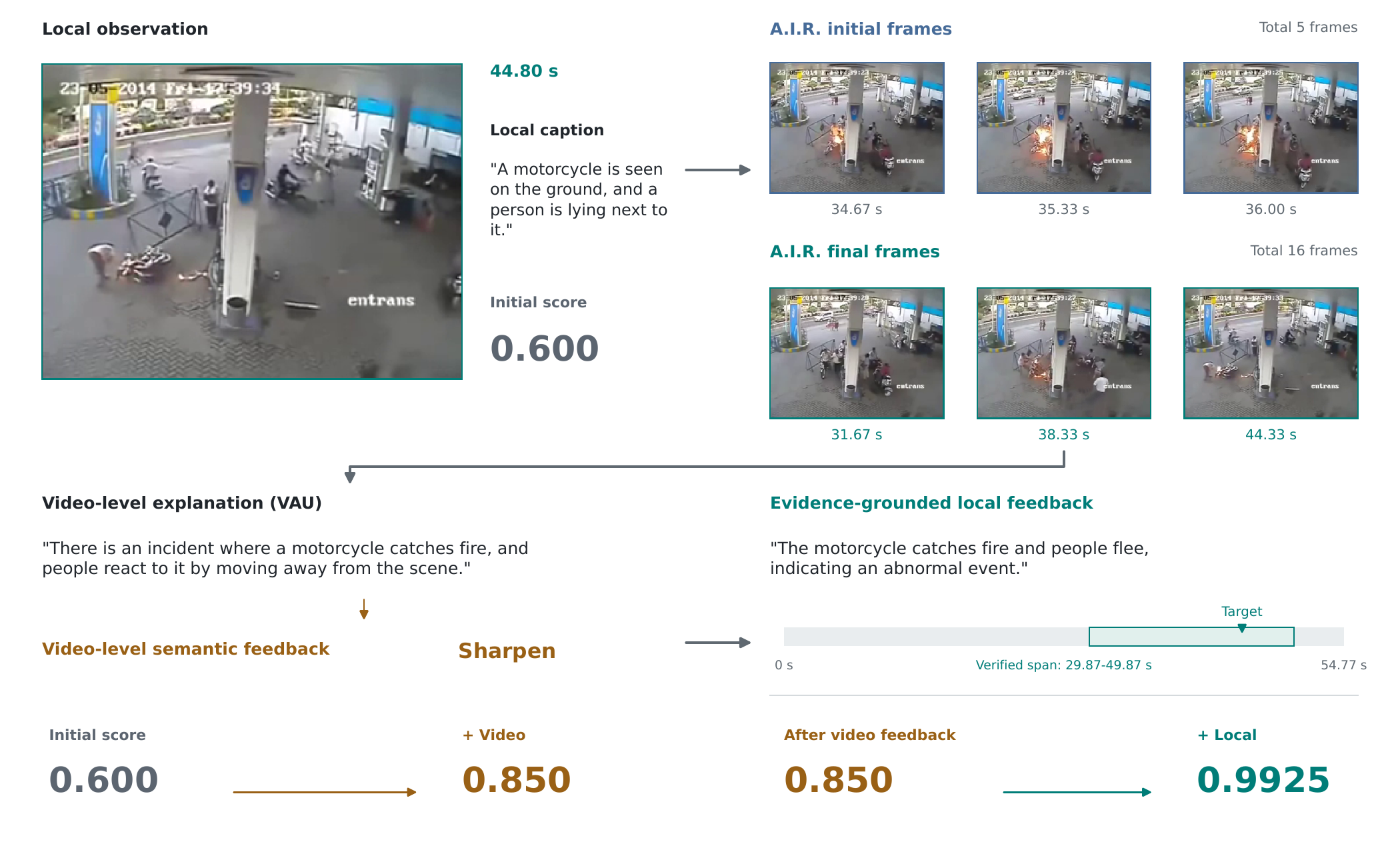}
    \caption{
Motorcycle-fire example. VAD-guided observations support the VAU description,
which guides video-level sharpening and visually authorized local promotion.
The displayed scores track both feedback stages at the illustrated location.
}
    \label{fig:vau-qualitative}
\end{figure}
Figure~\ref{fig:vau-qualitative} illustrates the progression from
VAD-guided observation to semantic feedback in a motorcycle-fire example.
The displayed local caption describes a motorcycle and a person on the
ground, while the final VAU description identifies a fire and people
moving away from the scene.
The description guides video-level sharpening, followed by visually
authorized local promotion. At the illustrated location, the score increases from
$0.600$ to $0.850$ after video-level feedback and to $0.9925$ after
local feedback.

\subsection{Ablation Studies}
\label{sec:mechanisms}

\begin{table*}[!ht]
\centering
\caption{
VAU observation ablation. All rows share the 16-frame answerer input budget,
answerer, prompting, spatial conditioning, and generation settings.
ATS and Ours share the VAD prior; AIR--CLIP and Ours share the active-observation
pipeline. Active observation additionally uses candidate-analysis calls.
GPT-Mean averages GPT-R, GPT-D, and GPT-C. Higher is better.
}
\label{tab:vau-ablation}

\scriptsize
\setlength{\tabcolsep}{2.15pt}
\renewcommand{\arraystretch}{1.12}

\resizebox{\textwidth}{!}{%
\begin{tabular}{@{}lcc|ccccc|ccccc@{}}
\toprule
\multirow{2}{*}{Method}
& \multirow{2}{*}{Guidance}
& \multirow{2}{*}{Active}
& \multicolumn{5}{c|}{UCF-Crime}
& \multicolumn{5}{c}{XD-Violence} \\
\cmidrule(lr){4-8}
\cmidrule(l){9-13}
& & &
BLEU & METEOR & ROUGE-L & CIDEr & GPT-Mean
& BLEU & METEOR & ROUGE-L & CIDEr & GPT-Mean \\
\midrule

ATS
& VAD & \no
& 0.3854 & 0.1829 & \textbf{0.1970} & \textbf{0.0357} & 0.4812
& 0.4344 & 0.1977 & 0.2089 & \textbf{0.0488} & 0.6724 \\

AIR--CLIP
& CLIP & \yes
& 0.3541 & 0.1735 & 0.1897 & 0.0230 & 0.4327
& 0.4334 & 0.2035 & 0.2068 & 0.0320 & 0.6897 \\

\textbf{Ours}
& VAD & \yes
& \textbf{0.3856} & \textbf{0.1833} & 0.1959 & 0.0232 & \textbf{0.4884}
& \textbf{0.4647} & \textbf{0.2111} & \textbf{0.2117} & 0.0468 & \textbf{0.7144} \\

\bottomrule
\end{tabular}%
}
\end{table*}

\paragraph{Anomaly guidance and active observation.}
Table~\ref{tab:vau-ablation} provides two controlled comparisons. The one-shot baseline is adapted from Holmes-VAU's
Anomaly-focused Temporal Sampler (ATS)~\citep{zhang2025holmes},
which samples frames according to the cumulative anomaly-score
distribution.
In our comparison, the baseline uses the same VAD prior and
16-frame final-generation conditions as our method, without
updating its selection from intermediate visual observations.
Within the active-observation pipeline, replacing CLIP guidance
with VAD guidance raises GPT-Mean by 0.0557 on UCF-Crime and
0.0247 on XD-Violence, supporting anomaly-score guidance for
the broad VAU instruction.

With the VAD prior and final-generation settings fixed, active
observation raises GPT-Mean over ATS by 0.0072 on UCF-Crime
and 0.0420 on XD-Violence. Lexical metrics are mixed on UCF-Crime.
On XD-Violence, BLEU, METEOR, and ROUGE-L improve, while CIDEr
remains slightly higher with ATS.

\begin{table}[!t]
\centering
\caption{
Cumulative ablation of the two semantic-feedback levels.
All rows use the URF-refined score base; feedback rows use the
VAU description generated by our VAD-guided active observation.
Video-level feedback includes selective visual rechecking, and the final row
additionally applies evidence-grounded local promotion.
}
\label{tab:vad-ablation}

\small
\setlength{\tabcolsep}{4.5pt}
\renewcommand{\arraystretch}{1.14}

\begin{tabular}{@{}lcccc@{}}
\toprule
& \multicolumn{2}{c}{UCF-Crime}
& \multicolumn{2}{c}{XD-Violence} \\
\cmidrule(lr){2-3}
\cmidrule(l){4-5}
Configuration
& AUC(\%) & AP(\%)
& AUC(\%) & AP(\%) \\
\midrule

URF-refined base
& 84.36 & 36.16
& 91.32 & 68.07 \\

+ Video-level feedback
& 86.23 & 37.57
& 91.64 & 71.37 \\

+ Local feedback (\textbf{Ours})
& \textbf{86.25} & \textbf{38.36}
& \textbf{91.74} & \textbf{72.58} \\

\bottomrule
\end{tabular}
\end{table}

\paragraph{Two-level semantic feedback.}
Table~\ref{tab:vad-ablation} shows that the video-level stage
yields the larger incremental gain on all four metrics in this
cumulative ablation. Local feedback provides further improvements:
0.79 AP points on UCF-Crime and 1.21 AP points on XD-Violence,
alongside AUC gains of 0.02 and 0.10 points, respectively.
Its additional contribution is thus more apparent in AP than
in AUC after video-level modulation.

\begin{table}[t]
\centering
\small
\caption{Mean scores and incremental changes under sequential refinement on UCF-Crime. GT-normal and GT-anomaly denote normal and anomalous temporal regions according to
the ground-truth annotations. Parentheses report changes from the preceding row.}
\label{tab:score-contrast}
\setlength{\tabcolsep}{4pt}
\renewcommand{\arraystretch}{1.14}
\begin{tabular}{@{}lr@{\hspace{0.35em}}lr@{\hspace{0.35em}}lr@{}}
\toprule
Score sequence & \multicolumn{2}{c}{GT-normal ($\Delta$)} & \multicolumn{2}{c}{GT-anomaly ($\Delta$)} & Difference \\
\midrule
RAW & 0.2217 & (--) & 0.3013 & (--) & 0.0796 \\
+URF-HVAA~\citep{lin2025urf} & 0.2445 & (+0.02286) & 0.3223 & (+0.02105) & 0.0778 \\
+Ours & 0.2449 & (+0.00036) & 0.3366 & (+0.01428) & 0.0917 \\
\bottomrule
\end{tabular}
\end{table}

\paragraph{Score changes in normal and anomalous regions.}
Table~\ref{tab:score-contrast} tracks mean scores from the RAW baseline
through URF-HVAA refinement and our semantic feedback.
URF-HVAA refinement produces similar mean increases in normal and anomalous regions (+0.02286 and +0.02105, respectively).
Our semantic feedback further increases the anomalous-region mean by +0.01428, while leaving the normal-region mean nearly unchanged (+0.00036).
Consequently, the mean within-video anomaly-background score difference increases from 0.0778 to 0.0917.
This more differentiated score response supports the effectiveness of considering which temporal regions warrant refinement when applying semantic feedback.

\paragraph{Parameter sensitivity.}
Appendix~\ref{app:feedback} evaluates each parameter while holding
the others fixed.
Semantic feedback improves both AUC and AP on both datasets across
all tested values of $\alpha$, $f$, and $\lambda$, and across multiple
settings of $\beta$, $m$, and the quantile thresholds.
The shared configuration balances AUC and AP gains across the two
datasets while maintaining selective score updates.
These results indicate that the improvements are not confined to
a single parameter setting.

\paragraph{Description source and refinement base.}
Appendix~\ref{app:additional} examines the effect of the description
source under a fixed feedback configuration.
With the URF-refined base and both feedback stages, our descriptions
yield higher AP than the released URF-HVAA descriptions on UCF-Crime
(38.36 vs.\ 38.16) and XD-Violence (72.58 vs.\ 71.49).
This comparison supports their use for detector refinement,
complementing the independent VAU evaluation.

\FloatBarrier
\section{Conclusion}
\label{sec:conclusion}

We presented a training-free framework that returns open-ended
anomaly understanding to temporal detection.
Its two feedback levels separate video-wide score modulation
from local promotion supported by target-clip evidence.
Controlled comparisons show that VAD-guided active observation
improves GPT-rated description quality, while cumulative
ablations show complementary gains from the two feedback stages.
These results support using a VAU description both as an
explanation and as a source of hypotheses for refining an
existing anomaly-score sequence.
\subsection*{AI Use Statement}

Generative AI tools were used to assist with literature retrieval, critical analysis of the proposed methodology, experimental analysis and interpretation, and the drafting and revision of portions of the manuscript. The motivation, problem formulation, and principal methodological ideas of this work were developed by the authors. AI-assisted analysis of experimental outcomes was also used to support subsequent methodological refinements, including the exploration and adjustment of experimental parameters and implementation choices. All AI-assisted suggestions, analyses, and generated text were reviewed and verified by the authors, who take full responsibility for the methodology, experiments, results, claims, and final content of the paper.
\FloatBarrier
\clearpage
\bibliographystyle{iclr2027_conference}
\bibliography{iclr2027_conference}
\clearpage
\appendix
\section{Additional analysis of semantic feedback}
\label{app:additional}

We additionally examine how semantic feedback behaves under different
description sources and refinement bases.
We compare the released URF-HVAA description with the description generated
by our VAD-guided active observation, while applying one shared feedback configuration across both datasets
and all compared variants.
We also evaluate both the original RAW score sequence and the URF-refined
score sequence as refinement bases.
Table~\ref{tab:providers} reports these variants as additional analysis
rather than as separate main methods.

\begin{table*}[h]
\centering
\caption{
Additional analysis of description sources, refinement bases, and feedback
stages.
``Video'' denotes video-level semantic feedback including selective visual
rechecking; ``Video + local'' additionally applies evidence-grounded local
promotion.
The same shared feedback configuration is used across datasets,
description sources, and refinement bases.
}
\label{tab:providers}

\scriptsize
\setlength{\tabcolsep}{4pt}
\renewcommand{\arraystretch}{1.13}

\begin{tabular}{@{}ll|cc|cc@{}}
\toprule
\multirow{2}{*}{Refinement base} & \multirow{2}{*}{Description / feedback}
& \multicolumn{2}{c|}{UCF-Crime} & \multicolumn{2}{c}{XD-Violence} \\
\cmidrule(lr){3-4}\cmidrule(l){5-6}
& & AUC & AP & AUC & AP \\
\midrule

\multirow{4}{*}{RAW} 
& URF-HVAA / Video 
& 84.33 & 35.73 & 88.78 & 64.26 \\

& URF-HVAA / Video + local 
& 84.31 & 35.92 & 89.23 & 66.17 \\

& Ours / Video 
& 84.33 & 36.36 & 88.95 & 64.69 \\

& Ours / Video + local 
& \textbf{84.35} & \textbf{37.10} & \textbf{89.29} & \textbf{66.61} \\

\midrule

\multirow{4}{*}{URF-refined} 
& URF-HVAA / Video 
& 86.21 & 37.61 & 91.55 & 70.63 \\

& URF-HVAA / Video + local 
& 86.19 & 38.16 & 91.61 & 71.49 \\

& Ours / Video 
& 86.23 & 37.57 & 91.64 & 71.37 \\

& Ours / Video + local 
& \textbf{86.25} & \textbf{38.36} & \textbf{91.74} & \textbf{72.58} \\
\bottomrule
\end{tabular}
\end{table*}

\paragraph{Description source and refinement base.}
Across both refinement bases, adding local feedback consistently improves AP
for all evaluated description sources and datasets.
On XD-Violence, local feedback also improves AUC in every configuration.
On UCF-Crime, the AUC effect is small and mixed: descriptions from our VAD-guided active observation
yield slight improvements, whereas URF-HVAA descriptions produce decreases
of only 0.02 points.
With both video-level and local feedback enabled, descriptions from our
VAD-guided active observation yield higher AP than URF-HVAA descriptions
for both refinement bases on UCF-Crime and XD-Violence.
The more consistent gains in AP than in AUC indicate that local
promotion benefits precision--recall performance in these comparisons,
although the aggregate metrics do not directly measure accuracy
within the retrieved windows.

\section{Observation and fixed-budget construction}
\label{app:observation}
\paragraph{Anomaly prior and initialization.}
We Gaussian-smooth $b^{(0)}$ with $\sigma=10$ and expand the
scores to frame level at stride 16, without additional
min--max normalization.
A two-component Gaussian mixture identifies high-score regions
from which the initial candidate frames are selected.

\paragraph{Active observation and final selection.}
We adapt the iterative observation procedure of
A.I.R.~\citep{zou2026air}, using anomaly scores to guide
initialization and interval prioritization.
The frozen candidate analyst assigns relevance scores from
1 to 5, which are combined with the anomaly scores to guide
subsequent candidate selection and localized resampling.
All runs use six observation rounds, with up to 24 candidate
indices per round and a nominal internal selection budget of
32 frames.

For the final answerer, we construct a 16-frame input using
anomaly and analyst scores with a temporal spacing preference.
Any deficit is filled with unused indices from a uniform
temporal grid, and the final frames are ordered chronologically.
The 16-frame budget applies to answer generation, not to the
total visual observations made during selection.
\paragraph{Interval potential.}
We retain the interval-ranking mechanism of
A.I.R.~\citep{zou2026air}.
For a temporal interval $I$, its core potential combines
relevance, temporal variation, and interval length:
\begin{equation}
P_{\mathrm{core}}(I)
=
\underbrace{R(I)}_{\text{relevance}}
\underbrace{\bigl(1+0.5C(I)\bigr)}_{\text{complexity}}
\underbrace{\bigl(1+0.2\log L_I\bigr)}_{\text{length}},
\end{equation}
where $R(I)$ aggregates relevance within the interval,
$C(I)$ measures its temporal variation, and $L_I$ is the
interval length in original-frame coordinates.
These factors favor relevant, temporally varying, and
longer intervals for further inspection.

Our adaptation replaces the underlying CLIP similarity
signal with the smoothed, frame-expanded scores derived
from $b^{(0)}$, while retaining the analyst's relevance
feedback.
Score aggregation and the additional exploration-density
adjustment follow the released A.I.R. implementation.
Endpoints of high-potential intervals provide candidates
for the next analyst evaluation.

\section{Visual Inputs and Generation Settings}
\label{app:generation}

\paragraph{Candidate analyst.}
Qwen2.5-VL-7B-Instruct assesses candidate frames independently
using the inherited relevance rubric, returning a score from
1 to 5 and an explanation.
Generation uses a 64-token output limit, temperature $0.01$,
top-$k=1$, and top-$p=0.0001$.

\paragraph{Final VAU generation.}
VideoLLaMA3-7B receives 16 selected frames in chronological order
with their timestamps.
We retain URF-HVAA's suspicious-phrase prompting and gated spatial
conditioning, using either original frames or the cached spatially
annotated source.
The final input contains neither analyst responses nor reference
descriptions.

The instruction requests anomalous activities and their explanation
without timestamps, or a normal-scene description if no anomaly
is found.
The answerer generates one description per video with a
1,024-token output limit, temperature $0.1$, and sampling enabled.
Compared observation conditions share the final-generation
settings and per-video seed.
The candidate and final-generation prompts are included with
the LaTeX source.
\section{Feedback Settings and Parameter Sensitivity}
\label{app:feedback}
\begin{table}[ht]
\centering
\caption{Feedback calibration parameters shared across all evaluated
datasets.}
\label{tab:parameters}
\small
\setlength{\tabcolsep}{4pt}
\begin{tabular}{@{}lrrrrrrr@{}}
\toprule
$\alpha$ & $\beta$ & $f$ & $m$
& Global quantile & $\lambda$ & $\tau_\ell$ \\
\midrule
0.40 & 0.80 & 0.85 & 0.45
& 0.70 & 0.95 & 0.85 \\
\bottomrule
\end{tabular}
\end{table}

\paragraph{Parameter sensitivity.}
Table~\ref{tab:parameters} gives the shared configuration, and
Tables~\ref{tab:sensitivity-normal-scale}--%
\ref{tab:sensitivity-span-quantile} report one-parameter
sensitivity analyses.
We evaluate one parameter at a time while fixing all remaining parameters
to the shared main configuration.
The ``Main config.'' row reports the performance obtained with the single
parameter setting used across all evaluated datasets.
For each sweep, the bold entry marks the value used in this shared
configuration.

Overall, the shared configuration lies in a relatively stable performance
region for most parameters rather than at an isolated per-metric optimum.
The suppression scale $\alpha$ is particularly stable around the selected
value, while $\beta$, $f$, and $\lambda$ mainly alter the balance between
AUC and AP.
The minimum peak threshold $m$, global quantile, and local quantile
$\tau_\ell$ exhibit clearer transition regions: overly permissive settings
can broaden score updates and substantially reduce performance.
Several individual sweep values outperform the main configuration on
particular metrics, indicating that the shared setting is not selected as
the independent optimum of each dataset or metric.

\paragraph{Video-level rechecking.}
The lexical interpretation distinguishes descriptions containing
normality expressions, anomaly expressions, both, or neither.
Visual rechecking is eligible when the description contains
anomaly expressions but at most $1\%$ of local captions do.
Qwen2-VL-7B-Instruct receives a 16-frame video overview and
12 frames from a 300-frame high-score target interval.
The textual input contains the VAU description and up to four
suspicious-category cues obtained by fixed rule matching of that
description. These cues do not use ground-truth categories.
Ambiguous or inconsistent verification preserves the input scores.

\paragraph{Local retrieval and verification.}
A separate parser extracts event hypotheses and enables local
promotion only when its overall judgment indicates an anomaly.
Caption windows use width 600 and stride 64 in original-frame
coordinates. TF--IDF cosine similarity ranks the windows, with
original-base score statistics resolving ties.
Up to three non-overlapping candidates are selected per event,
then filtered by a similarity threshold of $0.10$ and a minimum
original-base score of $0.50$ at caption anchors.
Rejected candidates are not replaced.

The verifier receives the event hypothesis, matched captions,
16 target frames, and eight context frames sampled over the
target interval extended by 300 frames on either side.
Promotion requires direct abnormal evidence within the target.
Approved windows use the sequential, current-score quantile
update defined in the main text.

\paragraph{Verifier generation.}
Both feedback stages use frozen Qwen2-VL-7B-Instruct with greedy
generation and output limits of 260 tokens for video-level checks
and 240 tokens for local checks.
Final score reconstruction and evaluation are described in
Appendix~\ref{app:metrics}.

\begin{table}[!htbp]
\centering
\small
\setlength{\tabcolsep}{4pt}
\caption{Sensitivity to $\alpha$ with all other calibration parameters fixed to the shared main configuration. Bold marks the value used in the main configuration. Parentheses report changes from the re-evaluated URF-refined base in percentage points.}
\label{tab:sensitivity-normal-scale}
\begin{tabular}{@{}l*{4}{r@{\hspace{0.35em}}l}@{}}
\toprule
$\alpha$ & \multicolumn{4}{c}{UCF-Crime} & \multicolumn{4}{c}{XD-Violence} \\
\cmidrule(lr){2-5}\cmidrule(lr){6-9}
 & \multicolumn{2}{c}{AUC (\%)} & \multicolumn{2}{c}{AP (\%)} & \multicolumn{2}{c}{AUC (\%)} & \multicolumn{2}{c}{AP (\%)} \\
\midrule
Baseline~\citep{lin2025urf} & 84.356 &  & 36.164 &  & 91.323 &  & 68.071 &  \\
Main config. & 86.246 & (+1.890) & 38.363 & (+2.199) & 91.738 & (+0.414) & 72.584 & (+4.514) \\
\midrule
0.05 & 86.551 & (+2.195) & 38.437 & (+2.273) & 91.656 & (+0.332) & 72.557 & (+4.486) \\
0.1 & 86.572 & (+2.216) & 38.439 & (+2.274) & 91.667 & (+0.343) & 72.559 & (+4.489) \\
0.15 & 86.587 & (+2.231) & 38.439 & (+2.275) & 91.680 & (+0.357) & 72.563 & (+4.492) \\
0.2 & 86.571 & (+2.215) & 38.435 & (+2.271) & 91.693 & (+0.370) & 72.567 & (+4.496) \\
0.3 & 86.461 & (+2.105) & 38.411 & (+2.247) & 91.718 & (+0.395) & 72.576 & (+4.505) \\
0.4 & \textbf{86.246} & \textbf{(+1.890)} & \textbf{38.363} & \textbf{(+2.199)} & \textbf{91.738} & \textbf{(+0.414)} & \textbf{72.584} & \textbf{(+4.514)} \\
0.425 & 86.184 & (+1.828) & 38.348 & (+2.184) & 91.742 & (+0.418) & 72.586 & (+4.516) \\
0.6 & 85.674 & (+1.318) & 38.212 & (+2.048) & 91.762 & (+0.438) & 72.597 & (+4.527) \\
0.8 & 84.975 & (+0.618) & 37.957 & (+1.793) & 91.767 & (+0.443) & 72.601 & (+4.530) \\
0.95 & 84.367 & (+0.010) & 37.649 & (+1.485) & 91.761 & (+0.438) & 72.597 & (+4.526) \\
\bottomrule
\end{tabular}
\end{table}

\begin{table}[!htbp]
\centering
\small
\setlength{\tabcolsep}{4pt}
\caption{Sensitivity to $\beta$ with all other calibration parameters fixed to the shared main configuration. Bold marks the value used in the main configuration. Parentheses report changes from the re-evaluated URF-refined base in percentage points.}
\label{tab:sensitivity-low-scale}
\begin{tabular}{@{}l*{4}{r@{\hspace{0.35em}}l}@{}}
\toprule
$\beta$ & \multicolumn{4}{c}{UCF-Crime} & \multicolumn{4}{c}{XD-Violence} \\
\cmidrule(lr){2-5}\cmidrule(lr){6-9}
 & \multicolumn{2}{c}{AUC (\%)} & \multicolumn{2}{c}{AP (\%)} & \multicolumn{2}{c}{AUC (\%)} & \multicolumn{2}{c}{AP (\%)} \\
\midrule
Baseline~\citep{lin2025urf} & 84.356 &  & 36.164 &  & 91.323 &  & 68.071 &  \\
Main config. & 86.246 & (+1.890) & 38.363 & (+2.199) & 91.738 & (+0.414) & 72.584 & (+4.514) \\
\midrule
0.1 & 82.938 & (-1.418) & 39.858 & (+3.694) & 90.113 & (-1.211) & 72.556 & (+4.485) \\
0.25 & 84.323 & (-0.033) & 39.973 & (+3.809) & 90.880 & (-0.443) & 73.199 & (+5.128) \\
0.4 & 85.251 & (+0.895) & 39.863 & (+3.699) & 91.354 & (+0.030) & 73.514 & (+5.444) \\
0.55 & 85.801 & (+1.445) & 39.513 & (+3.349) & 91.619 & (+0.296) & 73.492 & (+5.421) \\
0.7 & 86.121 & (+1.765) & 38.900 & (+2.736) & 91.731 & (+0.408) & 73.095 & (+5.024) \\
0.8 & \textbf{86.246} & \textbf{(+1.890)} & \textbf{38.363} & \textbf{(+2.199)} & \textbf{91.738} & \textbf{(+0.414)} & \textbf{72.584} & \textbf{(+4.514)} \\
0.9 & 86.324 & (+1.968) & 37.785 & (+1.620) & 91.699 & (+0.376) & 71.853 & (+3.782) \\
0.95 & 86.347 & (+1.991) & 37.467 & (+1.303) & 91.666 & (+0.342) & 71.388 & (+3.318) \\
\bottomrule
\end{tabular}
\end{table}

\begin{table}[!htbp]
\centering
\small
\setlength{\tabcolsep}{4pt}
\caption{Sensitivity to $f$ with all other calibration parameters fixed to the shared main configuration. Bold marks the value used in the main configuration. Parentheses report changes from the re-evaluated URF-refined base in percentage points.}
\label{tab:sensitivity-high-floor}
\begin{tabular}{@{}l*{4}{r@{\hspace{0.35em}}l}@{}}
\toprule
$f$ & \multicolumn{4}{c}{UCF-Crime} & \multicolumn{4}{c}{XD-Violence} \\
\cmidrule(lr){2-5}\cmidrule(lr){6-9}
 & \multicolumn{2}{c}{AUC (\%)} & \multicolumn{2}{c}{AP (\%)} & \multicolumn{2}{c}{AUC (\%)} & \multicolumn{2}{c}{AP (\%)} \\
\midrule
Baseline~\citep{lin2025urf} & 84.356 &  & 36.164 &  & 91.323 &  & 68.071 &  \\
Main config. & 86.246 & (+1.890) & 38.363 & (+2.199) & 91.738 & (+0.414) & 72.584 & (+4.514) \\
\midrule
0.5 & 86.197 & (+1.840) & 39.276 & (+3.111) & 91.522 & (+0.199) & 70.878 & (+2.807) \\
0.55 & 86.219 & (+1.863) & 39.319 & (+3.155) & 91.529 & (+0.206) & 70.882 & (+2.811) \\
0.65 & 86.267 & (+1.910) & 39.417 & (+3.252) & 91.575 & (+0.252) & 71.036 & (+2.965) \\
0.75 & 86.275 & (+1.919) & 39.367 & (+3.202) & 91.671 & (+0.348) & 71.521 & (+3.451) \\
0.85 & \textbf{86.246} & \textbf{(+1.890)} & \textbf{38.363} & \textbf{(+2.199)} & \textbf{91.738} & \textbf{(+0.414)} & \textbf{72.584} & \textbf{(+4.514)} \\
0.95 & 86.177 & (+1.820) & 36.940 & (+0.776) & 91.765 & (+0.442) & 72.704 & (+4.634) \\
1 & 86.134 & (+1.778) & 36.284 & (+0.120) & 91.755 & (+0.432) & 72.680 & (+4.610) \\
\bottomrule
\end{tabular}
\end{table}

\begin{table}[!htbp]
\centering
\small
\setlength{\tabcolsep}{4pt}
\caption{Sensitivity to $m$ with all other calibration parameters fixed to the shared main configuration. Bold marks the value used in the main configuration. Parentheses report changes from the re-evaluated URF-refined base in percentage points.}
\label{tab:sensitivity-peak-minimum}
\begin{tabular}{@{}l*{4}{r@{\hspace{0.35em}}l}@{}}
\toprule
$m$ & \multicolumn{4}{c}{UCF-Crime} & \multicolumn{4}{c}{XD-Violence} \\
\cmidrule(lr){2-5}\cmidrule(lr){6-9}
 & \multicolumn{2}{c}{AUC (\%)} & \multicolumn{2}{c}{AP (\%)} & \multicolumn{2}{c}{AUC (\%)} & \multicolumn{2}{c}{AP (\%)} \\
\midrule
Baseline~\citep{lin2025urf} & 84.356 &  & 36.164 &  & 91.323 &  & 68.071 &  \\
Main config. & 86.246 & (+1.890) & 38.363 & (+2.199) & 91.738 & (+0.414) & 72.584 & (+4.514) \\
\midrule
0.05 & 82.655 & (-1.701) & 26.925 & (-9.240) & 89.623 & (-1.700) & 68.973 & (+0.902) \\
0.1 & 82.655 & (-1.701) & 26.925 & (-9.240) & 89.623 & (-1.700) & 68.973 & (+0.902) \\
0.15 & 82.787 & (-1.569) & 27.005 & (-9.159) & 89.727 & (-1.596) & 69.090 & (+1.020) \\
0.2 & 82.787 & (-1.569) & 27.005 & (-9.159) & 89.727 & (-1.596) & 69.090 & (+1.020) \\
0.25 & 86.211 & (+1.855) & 37.045 & (+0.881) & 91.428 & (+0.105) & 72.120 & (+4.049) \\
0.35 & 86.341 & (+1.984) & 38.711 & (+2.547) & 91.662 & (+0.339) & 72.462 & (+4.391) \\
0.45 & \textbf{86.246} & \textbf{(+1.890)} & \textbf{38.363} & \textbf{(+2.199)} & \textbf{91.738} & \textbf{(+0.414)} & \textbf{72.584} & \textbf{(+4.514)} \\
0.55 & 86.092 & (+1.735) & 38.410 & (+2.246) & 91.703 & (+0.380) & 72.639 & (+4.568) \\
0.6 & 86.092 & (+1.735) & 38.410 & (+2.246) & 91.703 & (+0.380) & 72.639 & (+4.568) \\
\bottomrule
\end{tabular}
\end{table}

\begin{table}[!htbp]
\centering
\small
\setlength{\tabcolsep}{4pt}
\caption{Sensitivity to the global quantile with all other calibration parameters fixed to the shared main configuration. Bold marks the value used in the main configuration. Parentheses report changes from the re-evaluated URF-refined base in percentage points.}
\label{tab:sensitivity-peak-quantile}
\begin{tabular}{@{}l*{4}{r@{\hspace{0.35em}}l}@{}}
\toprule
Global quantile & \multicolumn{4}{c}{UCF-Crime} & \multicolumn{4}{c}{XD-Violence} \\
\cmidrule(lr){2-5}\cmidrule(lr){6-9}
 & \multicolumn{2}{c}{AUC (\%)} & \multicolumn{2}{c}{AP (\%)} & \multicolumn{2}{c}{AUC (\%)} & \multicolumn{2}{c}{AP (\%)} \\
\midrule
Baseline~\citep{lin2025urf} & 84.356 &  & 36.164 &  & 91.323 &  & 68.071 &  \\
Main config. & 86.246 & (+1.890) & 38.363 & (+2.199) & 91.738 & (+0.414) & 72.584 & (+4.514) \\
\midrule
0.1 & 86.117 & (+1.761) & 34.583 & (-1.581) & 91.432 & (+0.108) & 70.496 & (+2.425) \\
0.25 & 86.117 & (+1.761) & 34.583 & (-1.581) & 91.553 & (+0.229) & 71.716 & (+3.645) \\
0.4 & 86.138 & (+1.782) & 35.209 & (-0.956) & 91.733 & (+0.410) & 72.589 & (+4.518) \\
0.55 & 86.172 & (+1.816) & 35.875 & (-0.289) & 91.788 & (+0.465) & 72.952 & (+4.881) \\
0.7 & \textbf{86.246} & \textbf{(+1.890)} & \textbf{38.363} & \textbf{(+2.199)} & \textbf{91.738} & \textbf{(+0.414)} & \textbf{72.584} & \textbf{(+4.514)} \\
0.85 & 86.317 & (+1.961) & 39.755 & (+3.591) & 91.487 & (+0.163) & 70.661 & (+2.591) \\
0.95 & 86.386 & (+2.030) & 40.335 & (+4.171) & 91.484 & (+0.161) & 70.718 & (+2.647) \\
\bottomrule
\end{tabular}
\end{table}

\begin{table}[!htbp]
\centering
\small
\setlength{\tabcolsep}{4pt}
\caption{Sensitivity to $\lambda$ with all other calibration parameters fixed to the shared main configuration. Bold marks the value used in the main configuration. Parentheses report changes from the re-evaluated URF-refined base in percentage points.}
\label{tab:sensitivity-span-strength}
\begin{tabular}{@{}l*{4}{r@{\hspace{0.35em}}l}@{}}
\toprule
$\lambda$ & \multicolumn{4}{c}{UCF-Crime} & \multicolumn{4}{c}{XD-Violence} \\
\cmidrule(lr){2-5}\cmidrule(lr){6-9}
 & \multicolumn{2}{c}{AUC (\%)} & \multicolumn{2}{c}{AP (\%)} & \multicolumn{2}{c}{AUC (\%)} & \multicolumn{2}{c}{AP (\%)} \\
\midrule
Baseline~\citep{lin2025urf} & 84.356 &  & 36.164 &  & 91.323 &  & 68.071 &  \\
Main config. & 86.246 & (+1.890) & 38.363 & (+2.199) & 91.738 & (+0.414) & 72.584 & (+4.514) \\
\midrule
0.05 & 86.234 & (+1.877) & 37.654 & (+1.489) & 91.655 & (+0.332) & 71.460 & (+3.389) \\
0.1 & 86.241 & (+1.885) & 37.738 & (+1.573) & 91.667 & (+0.344) & 71.553 & (+3.482) \\
0.2 & 86.251 & (+1.895) & 37.862 & (+1.698) & 91.688 & (+0.365) & 71.734 & (+3.663) \\
0.3 & 86.257 & (+1.901) & 37.966 & (+1.802) & 91.704 & (+0.381) & 71.893 & (+3.823) \\
0.45 & 86.261 & (+1.905) & 38.113 & (+1.949) & 91.721 & (+0.398) & 72.114 & (+4.043) \\
0.65 & 86.259 & (+1.902) & 38.225 & (+2.061) & 91.734 & (+0.410) & 72.358 & (+4.287) \\
0.8 & 86.253 & (+1.897) & 38.298 & (+2.133) & 91.738 & (+0.415) & 72.488 & (+4.417) \\
0.95 & \textbf{86.246} & \textbf{(+1.890)} & \textbf{38.363} & \textbf{(+2.199)} & \textbf{91.738} & \textbf{(+0.414)} & \textbf{72.584} & \textbf{(+4.514)} \\
\bottomrule
\end{tabular}
\end{table}

\begin{table}[!htbp]
\centering
\small
\setlength{\tabcolsep}{4pt}
\caption{Sensitivity to $\tau_\ell$ with all other calibration parameters fixed to the shared main configuration. Bold marks the value used in the main configuration. Parentheses report changes from the re-evaluated URF-refined base in percentage points.}
\label{tab:sensitivity-span-quantile}
\begin{tabular}{@{}l*{4}{r@{\hspace{0.35em}}l}@{}}
\toprule
$\tau_\ell$ & \multicolumn{4}{c}{UCF-Crime} & \multicolumn{4}{c}{XD-Violence} \\
\cmidrule(lr){2-5}\cmidrule(lr){6-9}
 & \multicolumn{2}{c}{AUC (\%)} & \multicolumn{2}{c}{AP (\%)} & \multicolumn{2}{c}{AUC (\%)} & \multicolumn{2}{c}{AP (\%)} \\
\midrule
Baseline~\citep{lin2025urf} & 84.356 &  & 36.164 &  & 91.323 &  & 68.071 &  \\
Main config. & 86.246 & (+1.890) & 38.363 & (+2.199) & 91.738 & (+0.414) & 72.584 & (+4.514) \\
\midrule
0.1 & 85.889 & (+1.533) & 34.835 & (-1.330) & 89.747 & (-1.576) & 61.320 & (-6.751) \\
0.25 & 85.884 & (+1.527) & 33.707 & (-2.457) & 90.214 & (-1.109) & 63.830 & (-4.241) \\
0.4 & 86.018 & (+1.662) & 34.179 & (-1.985) & 90.962 & (-0.361) & 67.271 & (-0.800) \\
0.5 & 85.984 & (+1.628) & 34.666 & (-1.499) & 91.239 & (-0.084) & 68.766 & (+0.695) \\
0.65 & 86.141 & (+1.784) & 35.987 & (-0.177) & 91.789 & (+0.466) & 72.337 & (+4.266) \\
0.75 & 86.265 & (+1.909) & 37.829 & (+1.664) & 91.750 & (+0.427) & 72.274 & (+4.203) \\
0.85 & \textbf{86.246} & \textbf{(+1.890)} & \textbf{38.363} & \textbf{(+2.199)} & \textbf{91.738} & \textbf{(+0.414)} & \textbf{72.584} & \textbf{(+4.514)} \\
0.9 & 86.227 & (+1.871) & 37.840 & (+1.676) & 91.740 & (+0.417) & 72.473 & (+4.403) \\
\bottomrule
\end{tabular}
\end{table}

\section{Metric definitions and inference cost}
\label{app:metrics}
\paragraph{Lexical evaluation.}
We retain the metric labels and column order of the supplied URF-HVAA evaluation. In that implementation, the reported BLEU value is
\begin{equation}
\mathrm{BLEU}_{\mathrm{reported}}
=\frac{1}{n}\sum_{j=1}^{n}\sum_{k=1}^{4}
\mathrm{SentenceBLEU}_{k}(\hat y_j,y_j).
\end{equation}
Here $n$ is the number of evaluated videos, and $\hat{y}_j$ and $y_j$ denote the generated and reference descriptions for video $j$, respectively. The reported value sums sentence BLEU-1 through BLEU-4, rather than reporting BLEU-4 alone. NLTK smoothing method 4 and whitespace tokenization are used, with BLEU-3 weights $(0.33,0.33,0.33,0)$. METEOR uses NLTK \texttt{single\_meteor\_score}; ROUGE is ROUGE-L F-measure with stemming; CIDEr uses \texttt{pycocoevalcap}. Each dataset table uses its own CIDEr document-frequency corpus. Published reference values and re-evaluated released descriptions remain in separate blocks because they need not share identical evaluator outputs.

\paragraph{GPT evaluation.}
GPT evaluation uses \texttt{gpt-4.1-2025-04-14} at temperature zero.
Reasonability, detail, and consistency are assessed separately
from the generated and reference descriptions, without video input,
and averaged over each dataset.

\paragraph{Detection evaluation.}
After video and local updates, each score-anchor sequence is Gaussian-smoothed with $\sigma=10$, repeated at stride 16, truncated or zero-padded to the evaluation length, and concatenated with the corresponding frame labels. AUC and AP are computed from the pooled frame sequence with \texttt{roc\_auc\_score} and \texttt{average\_precision\_score}. AP is not a trapezoidal precision--recall integral. Neither per-video min--max normalization nor averaging of video-wise AUC is applied.

\paragraph{VAU inference cost.}
The analyst makes 18.22 image judgments per video on average,
and the final answerer receives 16 frames.
Recorded selection time averages 9.34 seconds per video.
Final generation takes 4.51 seconds within a 6.23-second
final-stage wall time.
These cache-based RTX 3090-class measurements exclude upstream
caption, tag, and spatial-annotation generation, model loading,
and feedback; they do not represent end-to-end latency.

\paragraph{Incremental VAD feedback cost.}
With scores, local captions, and the required descriptions or tags
already available, our feedback pipeline is $6.22\times$ faster than
full caption rescoring.
On the full UCF-Crime test set, Ours takes 743.12 seconds
(0.000668 seconds per frame), compared with 4622.26 seconds
for full caption rescoring, using two RTX 4090 GPUs.
Ours starts from URF-refined scores, while the control starts from RAW.
Timing includes preparation, decoding, fresh inference, score updates,
and I/O, but excludes upstream generation, model loading, warm-up,
final smoothing and expansion, and evaluation.
These single-run measurements use an earlier feedback calibration
and demonstrate an incremental processing advantage with precomputed
inputs, rather than an end-to-end VAD speedup.

\end{document}